\documentclass[letterpaper, 10 pt, conference]{ieeeconf}  % Comment this line out if you need a4paper
\usepackage{graphicx}  % 基本图片包
\usepackage{cite}  % 或者使用 natbib: \usepackage{natbib}
\usepackage{amsmath}
\usepackage{multirow} 
\usepackage{caption}  % 在导言区添加
\usepackage{xcolor}

\usepackage{subcaption}
\usepackage{float}

\usepackage{booktabs}

\IEEEoverridecommandlockouts                              % This command is only needed if 
\usepackage{amssymb}  % assumes amsmath package installed

\title{\LARGE \bf
Audio-VLA: Adding Contact Audio Perception to Vision-Language-Action Model for Robotic Manipulation}

\author{
Xiangyi Wei$^{1}$, Haotian Zhang$^{2}$, Xinyi Cao$^{3}$, Siyu Xie$^{3}$, Weifeng Ge$^{4}$, Yang Li$^{1}$, Changbo Wang$^{2}$\\[3pt]
\small $^{1}$School of Computer Science and Technology, East China Normal University\\
\small $^{2}$School of Data Science and Engineering, East China Normal University\\
\small $^{3}$School of Software Engineering, East China Normal University\\
\small $^{4}$School of Computer Science, Fudan University
}

\begin{document}

\maketitle
\thispagestyle{empty}
\pagestyle{empty}

%%%%%%%%%%%%%%%%%%%%%%%%%%%%%%%%%%%%%%%%%%%%%%%%%%%%%%%%%%%%%%%%%%%%%%%%%%%%%%%%
\begin{abstract}

The Vision-Language-Action models (VLA) have achieved significant advances in robotic manipulation recently.
However, vision-only VLA models create fundamental limitations, particularly in perceiving interactive and manipulation dynamic processes.
This paper proposes Audio-VLA, a multimodal manipulation policy that leverages contact audio to perceive contact events and dynamic process feedback.
Audio-VLA overcomes the vision-only constraints of VLA models. Additionally, this paper introduces the Task Completion Rate (TCR) metric to systematically evaluate dynamic operational processes.
Audio-VLA employs pre-trained DINOv2 and SigLIP as visual encoders, AudioCLIP as the audio encoder, and Llama2 as the large language model backbone.
We apply LoRA fine-tuning to these pre-trained modules to achieve robust cross-modal understanding of both visual and acoustic inputs.
A multimodal projection layer aligns features from different modalities into the same feature space.
Moreover RLBench and LIBERO simulation environments are enhanced by adding collision-based audio generation to provide realistic sound feedback during object interactions.
Since current robotic manipulation evaluations focus on final outcomes rather than providing systematic assessment of dynamic operational processes, the proposed TCR metric measures how well robots perceive dynamic processes during manipulation, creating a more comprehensive evaluation metric.
Extensive experiments on LIBERO, RLBench, and two real-world tasks demonstrate Audio-VLA's superior performance over vision-only comparative methods, while the TCR metric effectively quantifies dynamic process perception capabilities.

\end{abstract}

%%%%%%%%%%%%%%%%%%%%%%%%%%%%%%%%%%%%%%%%%%%%%%%%%%%%%%%%%%%%%%%%%%%%%%%%%%%%%%%%
\section{INTRODUCTION}

Robotic manipulation has emerged as one of the most challenging domains in robotics, requiring sophisticated perception and control capabilities to interact effectively with dynamic environments~\cite{1.1RT2,1.2Palm,1.3flamingo}.
Recent advances in Vision-Language-Action (VLA) models have demonstrated remarkable progress in learning manipulation policies from multimodal demonstrations~\cite{1.4openvla,1.5pi0}.
VLA models show promise in bridging the gap between high-level task understanding and low-level motor control.
Large-scale vision-language pre-training~\cite{1.4openvla} enables VLA models to achieve generalizable manipulation capabilities across diverse scenarios.

However, current VLA methods exhibit a fundamental limitation as they rely exclusively on visual perception~\cite{deitke2020robothor,o2024openx,LIBERO,james2020rlbench,mees2022calvin}.
The vision-only sensory modality introduces critical vulnerabilities in real-world deployment scenarios~\cite{1.6touch, 1.6touch2}.
Visual perception cannot adequately capture the rich dynamic information inherent in manipulation tasks, such as contact events~\cite{liu2024maniwav} and interaction feedback~\cite{mejia2024hearing}, which are crucial for precise manipulation control~\cite{hao2025tla, zhang2025vtla}.
From the perspective of embodied cognition theory~\cite{1.3six}, the vision-only constraint is even more severe—effective manipulation behavior requires establishing a complete understanding of the physical world through multi-sensory collaborative perception~\cite{1.3gupta2021embodied}.
Embodied cognition posits~\cite{1.3flamingo} that intelligence is not achieved merely through vision and reasoning, but rather emerges through real-time interaction between the body and the environment.
Prior works~\cite{hao2025tla,zhang2025vtla,cheng2025omnivtla} have attempted to use tactile signals to compensate for VLA's limited perception of dynamic information such as contact events and interaction feedback, but tactile sensors are expensive and  difficult to acquire and implement.
More critically, the low sampling frequencies of tactile sensors~\cite{van2018slip} limit perception of tool-object interactions and manipulation dynamics.

\begin{figure}[t]
\vspace{-\topskip}
\centering
\includegraphics[width=0.45\textwidth]{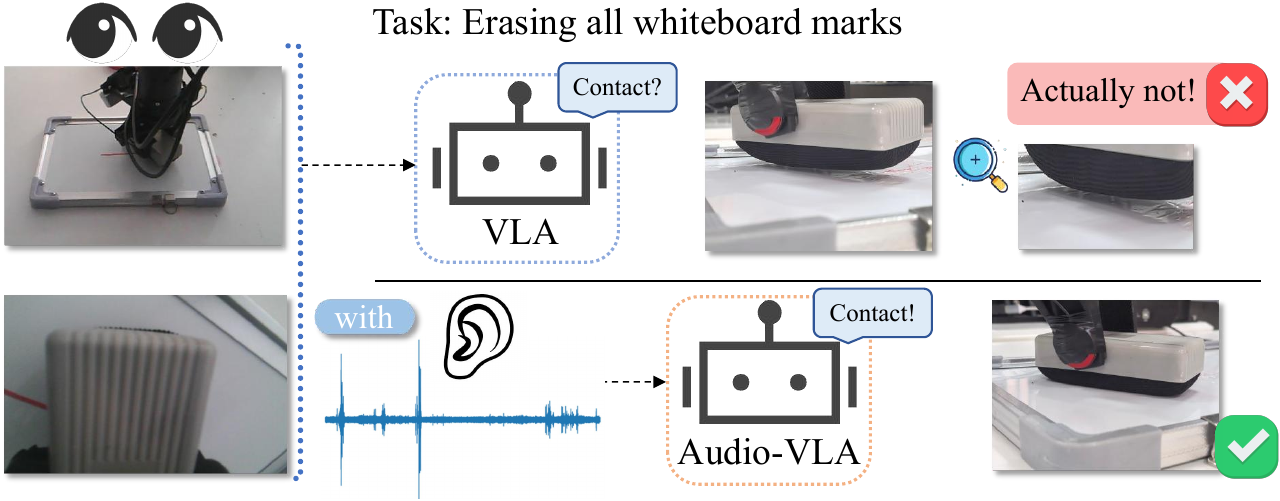}
\caption{Unlike  VLA models, Audio-VLA incorporates audio perception, enabling better assessment of contact states and understanding of manipulation dynamics.}
\label{fig:figure1}
\end{figure}

In contrast, contact audio~\cite{mejia2024hearing} offers a complementary perceptual modality providing rich information about object interactions, material properties, and dynamic processes during manipulation.
Unlike vision, audio signals penetrate occlusions and capture temporal dynamics that are visually imperceptible~\cite{liu2024maniwav,mejia2024hearing}.
As shown in Fig.~\ref{fig:figure1}, acoustic information reveals contact quality between objects and environmental changes during dynamic interactions—natural blind spots of visual perception that are indispensable for intelligent manipulation.
Contact audio has emerged as a promising alternative tactile sensing approach leveraging contact microphones to capture vibration signals from object interactions, providing real-time feedback about contact quality, material properties, and manipulation states~\cite{mejia2024hearing}.
Contact audio uniquely captures contact events at frequencies 1000× higher than conventional tactile sensors~\cite{ryun2017tactile}, while remaining immune to environmental variations like lighting and color changes that affect visual perception.
Furthermore, piezoelectric contact microphones provide an easily accessible, robust sensing solution~\cite{liu2024maniwav}.

Despite these promising capabilities, integrating acoustic information into existing VLA frameworks presents  technical challenges, such as extracting contact event information from high-frequency contact audio, and the lack of audio-enhanced training and evaluation environments.
To enable VLAs to perceive object interactions and dynamic processes during robotic manipulation.
In this paper, we propose Audio-VLA, a multimodal manipulation policy that combines acoustic and visual perception.
Our proposed Audio-VLA employs pre-trained Llama2~\cite{1.4llama2} as the backbone network, utilizes pre-trained DINOv2~\cite{oquab2023dinov2} and SigLIP~\cite{siglip} for visual encoder, and adopts AudioCLIP~\cite{guzhov2022audioclip} as audio encoder, with LoRA~\cite{hu2022lora} fine-tuning enabling robust cross-modal understanding.
A multi-modal projector is designed to map multi-modal features to text  feature space.
% Features from multiple modalities are mapped to the text feature space through multi-modal projector.
Additionally, to train and evaluate Audio-VLA in simulation environments and provide additional synthetic data, audio feedback based on collision detection is incorporated in both LIBERO~\cite{LIBERO} and RLBench~\cite{james2020rlbench} simulation environments. Furthermore, recognizing the limitations of existing evaluation metrics that focus primarily on final task outcomes, the Task Completion Rate (TCR) evaluation metric is proposed to quantify dynamic process perceptual feedback capabilities during manipulation execution providing comprehensive assessment of the policy's understanding of ongoing interaction dynamics rather than solely measuring end-state success.

Our proposed Audio-VLA achieves superior performance across both standard and domain shift~\cite{pumacay2024colosseum} LIBERO~\cite{LIBERO} and five RLBench~\cite{james2020rlbench} task benchmarks, particularly excelling in contact-intensive tasks. In real-world experiments, Audio-VLA demonstrates at least three-fold improvements in success rates compared to baseline methods under both seen and unseen environmental conditions, while also achieving the best TCR metrics. Extensive experiments  indicate that Audio-VLA can effectively extract contact event information from acoustic signals to overcome visual perception limitations, demonstrating robust dynamic process understanding capabilities.

\begin{figure*}[t]
\centering
\includegraphics[width=0.9\textwidth]{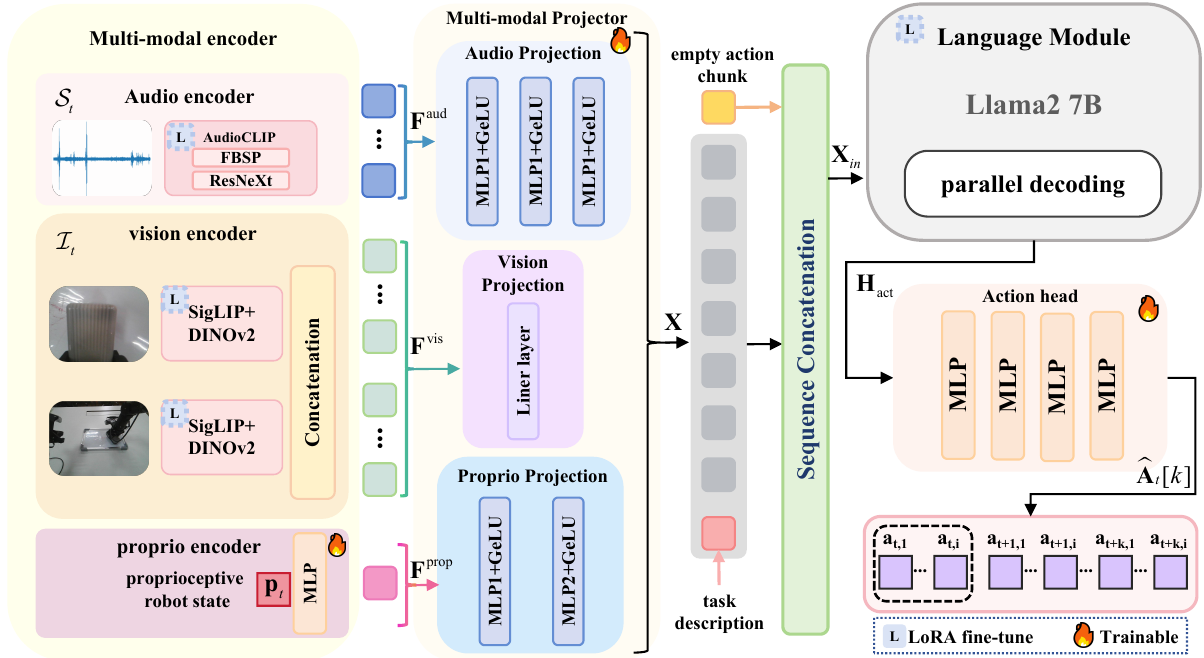}
\caption{Architecture of Audio-VLA. The model consists of multi-modal encoders including audio, vision, and proprioceptive modules, multi-modal Projector  that map heterogeneous features to a unified representation space, a 7B-parameter Llama2 language model as backbone, and a four-layer MLP action head for continuous action generation.}\label{fig:figure2}
\end{figure*}
\section{Related work}

\subsection{Vision-Language-Action models} 
The Robotic Transformer (RT-1)~\cite{rt1} pioneered transformer architectures for multi-task robotic manipulation, demonstrating generalizable skills across tasks by treating robot actions as language tokens. RT-2~\cite{1.1RT2} integrated vision-language models pre-trained on Internet-scale data with robotic demonstrations, achieving emergent capabilities like chain-of-thought reasoning and zero-shot generalization.
OpenVLA~\cite{1.4openvla} introduced the first open-source 7B-parameter VLA trained on 970,000 robot episodes, achieving competitive performance with proprietary systems. The $\pi0$ model~\cite{1.5pi0} explored continuous action generation through flow-matching networks, demonstrating superior performance for high-frequency dexterous control compared to discrete token-based approaches. CoT-VLA~\cite{cotvla} incorporated visual chain-of-thought reasoning by predicting future image frames as intermediate visual goals.
Despite these advances, existing VLA models rely exclusively on visual perception, limiting performance in robotic manipulation. Visual modality alone cannot capture critical contact dynamics and material interactions essential for contact-rich manipulation. Our work addresses this by integrating contact audio sensing into VLA frameworks, enabling perception of acoustic signatures during object interactions. This multimodal approach provides complementary information about contact quality and manipulation states beyond visual perception, enhancing VLA performance in complex manipulation scenarios.

\subsection{Audio and tactile enhanced robotic manipulation}
Hearing Touch~\cite{mejia2024hearing} and ManiWAV~\cite{liu2024maniwav} both use piezoelectric contact microphones to capture audio signals during manipulation. Hearing Touch~\cite{mejia2024hearing} leverages audio-visual pre-training to improve robotic performance, while ManiWAV proposes an “ear-in-hand" device for collecting contact sensing and learning policies from human demonstrations. While these works explore audio signals in robotic manipulation, they lack high-level semantic understanding. Our Audio-VLA integrates audio signals into the VLA framework, enabling complex natural language instruction processing with combined audio feedback.
Vision-based tactile sensors like GelSight~\cite{yuan2017gelsight} measure high-resolution geometry and force through optical deformation tracking at tens of microns resolution. Biomimetic sensors such as BioTac~\cite{narang2021interpreting} integrate force, vibration, and temperature to replicate human fingertip capabilities. While tactile integration shows promise~\cite{hao2025tla,zhang2025vtla,cheng2025omnivtla}, tactile sensors are costly, require end-effector integration, and only sense local contact. Our Audio-VLA captures acoustic signals through non-invasive contact microphones at low cost without mechanical alterations. Audio signals naturally capture temporal dynamics and environmental information beyond direct contact, providing richer contextual information for robotic manipulation.
%chen2022visuo

\section{Method}

Despite the success of VLA models, the lack of acoustic perception fundamentally limits their understanding of manipulation tasks from the perspective of environmental information extraction. We address this limitation through two key contributions: (1) Audio-VLA, a multimodal manipulation policy that integrates acoustic and visual perception, and (2) audio-enhanced simulation environments that enable large-scale training and evaluation. This section first details the Audio-VLA architecture, then presents our training objective and audio-enhanced simulation environments for LIBERO~\cite{LIBERO} and RLBench~\cite{james2020rlbench}.

\subsection{Architecture of Audio-VLA}

As shown in Fig.~\ref{fig:figure2}, the proposed Audio-VLA consists of four components including a multi-modal encoder, multi-modal projector, language module, and action head. 
Each component is introduced in detail below.

\paragraph{Multi-modal encoder} The multi-modal encoder comprises vision, audio, and proprio encoders.

The \textbf{vision encoder} processes raw images using a set of perceptual tokens. 
It consists of two powerful vision transformers, DINOv2~\cite{oquab2023dinov2} and SigLIP~\cite{siglip}, pre-trained on Internet-scale image data to capture rich visual features and comprehensive spatial understanding of the observations. 
To adapt these pre-trained encoders to the specific task while maintaining computational efficiency, LoRA~\cite{hu2022lora} is employed to fine-tune the vision encoder.
At each timestep $t$, the visual input is: % Added definition for timestep t
\begin{equation}
\mathcal{I}_t = \{ I_t^{\text{3rd}}, I_t^{\text{wrist}} \}
\end{equation}
where $\mathcal{I}_t$ denotes the visual input set at timestep $t$, $I^{\text{3rd}}_t$ is the third-person view image, and $I^{\text{wrist}}_t$ is the wrist camera image. % Added definitions for visual input symbols
$I_t^{(\cdot)} \in \mathbb{R}^{B \times 3 \times H \times W}$ denotes the raw RGB images from different camera viewpoints at timestep $t$, with $B$ representing the batch size, $H=W$ representing the image dimensions. % Added definitions for B, H, W
Visual features are extracted by SigLIP~\cite{siglip} and DINOv2~\cite{oquab2023dinov2}:
\begin{equation}
\mathbf{F}^{(\cdot)} = \mathcal{F}(I_t^{(\cdot)})
\end{equation}
where $\mathcal{F}$ represents the fused visual encoder, and $\mathbf{F}^{(\cdot)} \in \mathbb{R}^{H \times N_p \times G}$ denotes the extracted visual features from cameras, with $N_p = 256$ representing the number of visual patches and $G$ representing the patch embedding dimension. % Added definitions for visual encoder and feature dimensions
The visual features $\mathbf{F}^{\text{vis}}$ input to the language model are:
\begin{equation}
\mathbf{F}^{\text{vis}} = \text{Concat}(\mathbf{F}^{\text{3rd}}, \mathbf{F}^{\text{wrist}}) \in \mathbb{R}^{N_v \times d_{\text{vis}}}
\end{equation}
where $N_v$ denotes the total number of visual tokens and $d_{\text{vis}}$ represents the visual feature dimension. % Added definitions for concatenated visual features

The \textbf{audio encoder} employs AudioCLIP~\cite{guzhov2022audioclip}, a model pre-trained on large-scale audio-visual data~\cite{audioset,deng2009imagenet,clip} to learn rich audio representations. 
To enhance AudioCLIP's capability in perceiving robotic contact events, additional training is conducted on the ManiWAV~\cite{liu2024maniwav} robotic manipulation dataset based on the original pretrained weights, with LoRA~\cite{hu2022lora} fine-tuning further applied within Audio-VLA to optimize contact event detection performance.

Contact events generate high-frequency acoustic signals rich in physical interaction information. 
To fully leverage these signals, we optimize for high-frequency feature extraction and fine-grained temporal modeling. 
Unlike prior audio processing methods~\cite{alamri2019audio,mejia2024hearing}, we process audio signals independently at each timestep, enabling instantaneous contact event detection and precise temporal alignment with visual, proprioceptive, and action modalities. 
At each timestep $t$, the audio input is $\mathcal{S}_t \in \mathbb{R}^{B \times C \times T}$, where $C$ denotes the number of audio channels and $T$ represents the sampling points for every timestep $t$ temporal windows. % Added definitions for audio input dimensions
In the Frequency B-Spline Projection (FBSP) layer~\cite{guzhov2022audioclip}, we reduced both the hop length and window length to enhance temporal resolution while maintaining sufficient frequency resolution, thereby achieving finer-grained temporal modeling while preserving the effective capture capability for high-frequency contact events:
\begin{equation}
\mathbf{Z} = \text{FBSP}(\mathcal{S}_{t}, L_{win}, L_{hop})
\end{equation}
where $\mathbf{Z} \in \mathbb{R}^{B \times 1025 \times N_f}$ represents the complex-valued spectrogram, $L_{win} = 1024$ denotes the window length, $L_{hop} = 256$ indicates the hop length, and $N_f$ represents the number of frames. % Added definitions for FBSP parameters and output

Subsequently, the complex-valued spectrogram $\mathbf{Z}$ undergoes power spectrum computation, logarithmic scaling, and feature extraction through the ResNeXt~\cite{guzhov2022audioclip} layer:
\begin{equation}
\mathbf{F}^{\text{aud}} = \text{ResNeXt}(10 \cdot \log_{10}(|\mathbf{Z}|^2 + \epsilon))
\end{equation}
where $\mathbf{F}^{\text{aud}} \in \mathbb{R}^{N_a \times d_{\text{aud}}}$ represents the final audio feature embeddings, $N_a$ denotes the number of audio tokens, $d_{\text{aud}}$ represents the audio feature dimension, and $\epsilon = 10^{-18}$ prevents numerical instabilities during logarithmic computation. % Added definitions for audio features and epsilon
The complex-valued spectrogram $|\mathbf{Z}|^2$ extracts energy information across frequency components, while logarithmic scaling enhances dynamic range and compresses magnitude differences in acoustic signals. 
This enables $\mathbf{F}^{\text{aud}}$ to capture high-frequency acoustic features and temporal dynamics of contact events, providing physical interaction information unavailable through visual perception alone.

The input of the \textbf{proprio encoder} is proprioceptive robot state $\mathbf{p}_t$, which includes information such as joint angles, encoded via an MLP layer $\phi_{\text{state}}(\cdot)$ to obtain the state embedding: % Added "the" before "proprio module" for grammatical correctness
\begin{equation}
\mathbf{F}^{\text{prop}} = \phi_{\text{state}}(\mathbf{p}_t) \in \mathbb{R}^{1 \times d_{\text{prop}}}
\end{equation}
where $\mathbf{p}_t$ represents the proprioceptive robot state at timestep $t$,  and $d_{\text{prop}}$ represents the proprioceptive feature dimension. % Added definitions for proprioceptive symbols

\paragraph{Multi-Modal Projector} 

To enable the language model to process heterogeneous multi-modal inputs, specialized projection layers transform domain-specific features into the unified embedding space of the LLM. 
These projections preserve temporal structure during mapping, enabling the LLM to capture visual motion dynamics, time-varying audio information, and robot state evolution while modeling their temporal dependencies.

Three projection functions are defined: visual projection $\phi_v$ as a linear transformation, audio projection $\phi_a$ as a three-layer MLP, and proprioceptive projection $\phi_p$ as a two-layer MLP, each mapping the corresponding modality features to the unified representation space of dimension $d_{\text{llm}}$. % Added definition for LLM dimension

The complete multi-modal projector process constructs a unified temporal sequence by concatenating the projected features along the sequence dimension:
\begin{equation}
\mathbf{X} = [\phi_v(\mathbf{F}^{\text{vis}}); \phi_a(\mathbf{F}^{\text{aud}}); \phi_p(\mathbf{F}^{\text{prop}})] \in \mathbb{R}^{(N_v + N_a + 1) \times d_{\text{llm}}}
\end{equation}

This unified representation $\mathbf{X}$ preserves intra-modal temporal continuity and establishes cross-modal temporal alignment, enabling the LLM to leverage its sequence modeling capabilities to understand temporal evolution and causal relationships in multi-modal information.

\paragraph{Language Module} 

The language module performs cognitive reasoning by integrating multimodal sensory data with natural language commands. 
Built upon the 7B-parameter Llama2~\cite{1.4llama2} architecture, this module first converts the language instruction $l$ into a sequence of linguistic tokens through Llama2's tokenizer $\phi_{\text{emb}}$: % Removed braces around l for consistency
\begin{equation}
\mathbf{E}_{\text{lang}} = \phi_{\text{emb}}(\text{Tokenize}(l)) \in \mathbb{R}^{N_{l} \times d_{\text{llm}}}
\end{equation}
where $l$ represents the natural language instruction, $\phi_{\text{emb}}$ denotes the tokenizer embedding function, $\mathbf{E}_{\text{lang}}$ represents the embedded language tokens, and $N_{l}$ denotes the number of tokens in the language instruction. % Added definitions for language processing symbols

The tokens from all modalities are concatenated with $K \cdot D$ learnable empty action embeddings $\mathbf{E}_{\text{empty}} \in \mathbb{R}^{(K \cdot D) \times d_{\text{llm}}}$ to form the complete input sequence:
\begin{equation}
\mathbf{X}_{\text{in}} = [\mathbf{E}_{\text{lang}}; \mathbf{X}; \mathbf{E}_{\text{empty}}] \in \mathbb{R}^{(N_{l} + N_v + N_a + 1 + K \cdot D) \times d_{\text{llm}}}
\end{equation}
where $K$ denotes the number of future timesteps predicted at once, and $D$ denotes the dimensionality of a single-step control command of the robot. 
The complete input sequence $\mathbf{X}_{\text{in}}$ is fed into Llama2 for parallel decoding~\cite{1.4llama2} to obtain the hidden state sequence $\mathbf{H}_{\text{dec}}$. 
This process can be formally expressed as:
\begin{equation}
\mathbf{H}_{\text{dec}} = \text{Llama2}(\mathbf{X}_{\text{in}})
\end{equation}
where $\mathbf{H}_{\text{dec}}$ represents the decoded hidden states from the language model. % Added definition for decoded hidden states

Subsequently, we extract the action hidden states $\mathbf{H}_{\text{act}}$ from $\mathbf{H}_{\text{dec}}$, where each vector $\mathbf{h}^{(m)} \in \mathbb{R}^{d_{\text{llm}}}$ for $m = 1, \ldots, K \cdot D$ encodes contextual information from all input modalities for predicting a specific action dimension.

\paragraph{Action Head} 

The action head $\phi_{\text{act}}$ independently processes each vector in $\mathbf{H}_{\text{act}}$ to generate $K$ continuous actions, each of dimension $D$, corresponding to future timesteps. 
These predictions are subsequently reshaped into an action block $\hat{\mathbf{A}}_t$:
\begin{equation}
\hat{\mathbf{A}}_t = \text{Reshape}(\phi_{\text{act}}(\mathbf{H}_{\text{act}})) \in [-1, 1]^{K \times D}
\end{equation}
% where $\hat{\mathbf{A}}_t$ denotes the model-predicted action block comprising $K$ consecutive timesteps.
% Here, $\hat{\mathbf{a}}_{t+k} = \hat{\mathbf{A}}_t[k] \in [-1, 1]^D$ represents the predicted action vector at timestep $t+k$, where $k \in \{0, 1, \ldots, K-1\}$.
where $\hat{\mathbf{A}}_t$ denotes the model-predicted action block comprising $K$ consecutive timesteps, with each element $\hat{\mathbf{a}}_{t+k} = \hat{\mathbf{A}}_t[k] \in [-1, 1]^D$ representing the predicted action vector at timestep $t+k$ for $k \in \{0, 1, \ldots, K-1\}$.

\subsection{Training objective}

Our proposed Audio-VLA aims to minimize the discrepancy between the predicted action block $\hat{\mathbf{A}}_t$ and the expert-demonstrated ground truth action block $\mathbf{A}^*_t$. 
This is achieved through the minimization of the mean L1 loss function:
\begin{equation}
\mathcal{L} = \frac{1}{K \cdot D} \sum_{k=0}^{K-1} \sum_{i=1}^{D} \left| \hat{a}_{t+k,i} - a_{t+k,i}^* \right|
\end{equation}
where $\hat{a}_{t+k,i}$ denotes the $i$-th dimension value of the predicted action at timestep $t+k$, $a_{t+k,i}^*$ denotes the corresponding ground truth value, and $\mathcal{L}$ represents the loss function. % Simplified explanation and added loss function definition

\subsection{Audio-Enhanced Simulation Environments}
\label{audio-enhance}
To train and evaluate our proposed Audio-VLA, we augment LIBERO~\cite{LIBERO} and RLBench~\cite{james2020rlbench} simulation environments with realistic acoustic feedback via integrating real-world audio recordings triggered through physics-based collision detection.
Our approach systematically collects contact sounds from physical manipulation, for each simulated task, we identify target objects by material properties and dimensions, then perform equivalent manipulations on similar real-world objects while recording contact audio at 48kHz using gripper-mounted microphones.
These recordings are organized into a structured library indexed by material pairs, interaction types, and force magnitudes. 

The collected audio recordings are organized into a structured library indexed by material pairs, interaction types, and force magnitudes. 
During simulation, two types of collision events are monitored, including direct gripper–object contacts and interactions between grasped objects and the environment. When a collision is detected, the participants, impact velocity, and force magnitude are identified. The identified collision parameters are then used to query the audio library and retrieve appropriate sound samples. The retrieved audio is dynamically modulated, with amplitude scaled by collision force, pitch shifted according to object size, and duration adjusted for continuous contacts. Both audio-enhanced simulation environments will be made open-source to facilitate future research.

\section{Experiment}

This paper investigates acoustic perception in robotic manipulation through two questions: (1) \textbf{Is contact audio a critical signal for manipulation?} Whether acoustic feedback improves performance in scenarios where contact dynamics are difficult to perceive visually. (2) \textbf{Can contact audio enable dynamic process understanding?} Whether audio helps policies better comprehend ongoing interactions. To address these questions, experiments are conducted across simulation and real-world environments.

\subsection{Setup}
\paragraph{Simulation setup}
Simulation experiments are conducted on the audio-enhanced simulation environments introduced in Section.~\ref{audio-enhance}. Training is performed in the standard environments with incorporated audio feedback, following all official configurations. The training data consist of the official LIBERO dataset~\cite{LIBERO} and the RLBench demonstrations collected by Shridhar et al.~\cite{peract}, with 100 demonstrations provided for each RLBench task. Besides evaluation in standard environments identical to training settings, assessment is also conducted under domain shift conditions to evaluate generalization capabilities. Following Sharma et al.~\cite{pumacay2024colosseum}, domain shift is implemented by randomly varying lighting conditions and desktop surface material colors throughout the environments. 

\paragraph{Real-world setup}
The real-world experimental setup encompasses  robot platform, task design, data collection, and evaluation protocols.

\begin{figure}[t]
    \centering
    \begin{subfigure}[b]{0.25\textwidth}
        \includegraphics[width=\textwidth]{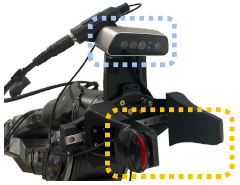}
        \caption{Piezo contact microphone and wrist camera installation}
        \label{fig:task2a}
    \end{subfigure}
    \hfill
    \begin{subfigure}[b]{0.22\textwidth}
        \includegraphics[width=\textwidth]{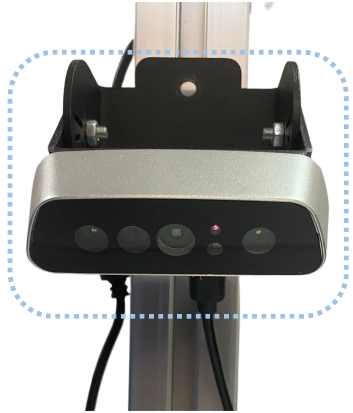}
        \caption{Third-person camera setup}
        \label{fig:task2b}
    \end{subfigure}
    
    \begin{subfigure}[b]{0.48\textwidth}
        \centering
        \includegraphics[width=\textwidth]{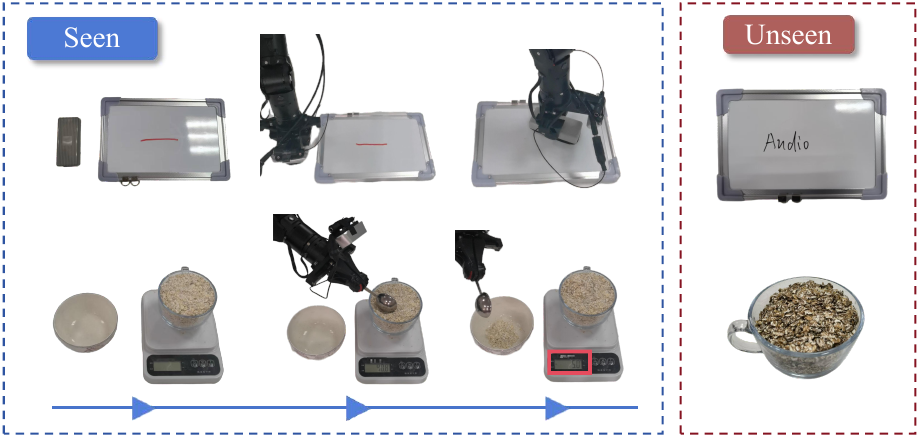}
        \caption{EAWM and S5GO. The first three columns represent training environments, and the last column represents the shift to unseen environment.}
        \label{fig:task2c}
    \end{subfigure}
    
    \caption{Experimental setup showing the hardware platform and real-world manipulation tasks}
    \label{fig:task2}
\end{figure}

\noindent\textbf{Platform:} 
Our experiments employ the AgileX Mobile ALOHA~\cite{fu2024mobile} platform with dual 7-DOF Piper robotic arms. We use the right arm equipped with wrist and third-person cameras from Orbbec dabaidc1 for visual perception and piezo contact microphones attached to the gripper for acoustic sensing, as shown in Fig.\ref{fig:task2a} and Fig.\ref{fig:task2b}.

\noindent\textbf{Real world Tasks:}
In order to provide comprehensive evaluation of acoustic perception in real-world manipulation scenarios, as shown in Fig.~\ref{fig:task2c}, two tasks with distinct contact dynamics are designed, each focusing on different aspects of contact audio feedback: \textbf{Erasing All Whiteboard Marks (EAWM)}  and \textbf{Scooping 5 Grams of Oatmeal (S5GO)}.

\noindent\textbf{Data Collections:}
Demonstrations are collected via teleoperation in a master-slave configuration. Visual data is captured at 480$\times$640 resolution, while two piezoelectric contact microphones mounted on both sides of the gripper record audio at 44.1kHz sampling rate. The raw audio stream is segmented into fixed-length chunks corresponding to each timestep for temporal alignment with visual observations. A total of 40 demonstrations are collected for each task.

\noindent\textbf{Evaluation Protocols:}
Evaluations are conducted with inference on an NVIDIA H20 GPU communicating with the robot platform through ROS~\cite{quigley2009ros}, where inference and execution run in parallel for continuous control at 25Hz. Real-world evaluations are conducted under seen and unseen conditions. Seen environments maintain identical conditions to training, while unseen environments introduce systematic variations to assess generalization.
For EAWM, a black marker creates different shapes on the whiteboard compared to training, and for S5GO, a darker oatmeal variety with distinct granular properties is employed.
In seen environments, for the EAWM task, we draw approximately 10-centimeter straight lines on the whiteboard using a red marker, with the objective of completely erasing these lines using an eraser. For the S5GO task, the objective is to scoop 5 grams of oatmeal from a bowl filled with oatmeal using a spoon, with the scooping process continuously monitored using a food scale.
Besides success rate, this work introduces TCR as a continuous metric measuring partial task achievements and providing insight into dynamic process understanding.
Formally, TCR is defined as:\begin{equation}
    \text{TCR} = \frac{\text{Achieved Progress}}{\text{Task Target}} \in [0, 1]
    \end{equation}
where the achieved progress is task-specific, in this paper:
    \begin{itemize}
        \item \textit{EAWM}: Area of erased marks
        \item \textit{S5GO}: Weight of scooped oatmeal
    \end{itemize}

% 详细数据表格
\begin{table*}[t]
\centering
\caption{Task success rates (\%) of Audio-VLA and comparative methods on LIBERO and RLBench benchmarks under standard environment and domain shift conditions. Tasks 1 to 5 correspond to close jar, insert onto square peg, light bulb in, open drawer, and put item in drawer, respectively.}
\label{tab:simulation_results}

% 插入生成的PDF图表

% 详细数据表格
\resizebox{\textwidth}{!}{%
\begin{tabular}{l|ccccc|cccccc}
\toprule
& \multicolumn{5}{c|}{\textbf{LIBERO}} & \multicolumn{6}{c}{\textbf{RLBench}} \\
\midrule
\textbf{Method} & Spatial & Object & Goal & Long & \textbf{Avg} & Task1 & Task2 & Task3 & Task4 & Task5 & \textbf{Avg} \\ 
\midrule
\multicolumn{12}{c}{\textit{Standard Environment}} \\
\midrule
$\pi_0$-FAST  & 96.4 & 96.8 & 88.8 & 60.2 & 85.6 & 62.1 & 8.9 & 5.2 & 75.0 & 68.3 & 43.9 \\
Cot-VLA & 87.5 & 91.6 & 87.6 & 69.0 & 83.9 & - & - & - & - & - & - \\
OpenVLA-OFT & 97.6 & 98.4 & 97.9 & 94.5 & 97.1 & 68.3 & 12.3 & 8.7 & 81.2 & 70.1 & 48.1 \\
\textbf{Audio-VLA (Ours)} & \textbf{98.3} & \textbf{98.5} & \textbf{98.2} & \textbf{95.4} & \textbf{97.6} & \textbf{75.1} & \textbf{15.0} & \textbf{18.9} & \textbf{87.5} & \textbf{79.2} & \textbf{55.1} \\ 
\midrule
\multicolumn{12}{c}{\textit{Domain shift}} \\
\midrule
$\pi_0$-FAST  & 72.3 & 72.6 & 66.6 & 45.2 & 64.2 & 46.6 & 4.0 & 2.2 & 56.3 & 51.2 & 32.1 \\
Cot-VLA & 65.6 & 68.7 & 65.7 & 31.1 & 57.8 & - & - & - & - & - & - \\
OpenVLA-OFT & 73.2 & 73.8 & 65.9 & 70.9 & 71.0 & 51.2 & 6.1 & 3.2 & 60.9 & 56.2 & 35.5 \\
\textbf{Audio-VLA (Ours)} & \textbf{75.7} & \textbf{76.9} & \textbf{73.7} & \textbf{72.3} & \textbf{74.7} & \textbf{56.3} & \textbf{10.0} & \textbf{13.2} & \textbf{68.6} & \textbf{59.4} & \textbf{41.5} \\ 
\bottomrule
\end{tabular}%
}
\end{table*}

\begin{table}[t]
\centering
\caption{Real-world manipulation task performance}
\begin{tabular}{ccccc}
\hline
\multirow{2}{*}{Method}   & \multicolumn{2}{c}{EAWS} & \multicolumn{2}{c}{S5GO} \\
                          & Success rate      & TCR      & Success rate         & TCR         \\ \hline
\multicolumn{5}{c}{\textit{Seen Environment}} \\
\hline
$\pi_0$-FAST              & 20                     & 34            & 10                     &    23          \\
OpenVLA-OFT               & 20                     & 45            & 10                     &    34          \\
\textbf{Audio-VLA (Ours)} & \textbf{60}           & \textbf{73}   & \textbf{30}            & \textbf{72}    \\ \hline
\multicolumn{5}{c}{\textit{Unseen Environment}} \\
\hline
$\pi_0$-FAST              &   0                   & 16             & 0                      &      11         \\
OpenVLA-OFT               & 10                      & 26             & 0                      &    24           \\
\textbf{Audio-VLA (Ours)} & \textbf{30}            & \textbf{57}    & \textbf{20}             & \textbf{56}     \\ \hline
\end{tabular}
\label{tab:real_world_results}
\end{table}

\begin{table}[t]
\centering
\caption{Ablation study results on RLBench}
\begin{tabular}{ccccccc}
\hline
Configuration & Task1 & Task2 & Task3 & Task4 & Task5 & Avg  \\ \hline
w/o LoRA      & 70.0  & 13.5  & 15.0  & 85.0  & 74.0  & 51.5 \\
Vision-only   & 68.0  & 12.0  & 9.0   & 82.0  & 69.0  & 48.0 \\
\textbf{Full}          & \textbf{75.1}  & \textbf{15}    & \textbf{18.9}  & \textbf{87.5}  & \textbf{79.2}  & \textbf{55.1} \\ \hline
\end{tabular}
\label{tab:ablation_sim_results}
\end{table}

\begin{table}[t]
\centering
\caption{Ablation study results on Real-world}
\begin{tabular}{ccccc}
\hline
\multirow{2}{*}{Configuration} & \multicolumn{2}{c}{EAWS} & \multicolumn{2}{c}{S5GO} \\
                               & Success rate       & TCR       & Success rate          & TCR          \\ \hline
       Vision-only                & 20                     & 46            & 10                     &   35           \\
   w/o LoRA                 & 30                     & 52          & 10                     &      46        \\
\textbf{Full}                  & \textbf{60}                     &    \textbf{73}       & \textbf{30}                     &     \textbf{72}         \\ \hline \end{tabular}
\label{tab:ablation_real_results}
\end{table}

\noindent\textbf{Implementations:}
For each task, all models are trained on 2× NVIDIA H20 GPUs.  The LoRA~\cite{hu2022lora} rank is set to 32, training runs for 50k to 100k steps depending on the task, the batch size is 8, and the learning rate is 1e-4 with cosine annealing.
Both Audio-VLA and comparative methods use identical training configurations to ensure fair comparison.

\subsection{Experimental Results}

We evaluate Audio-VLA against all comparative methods on the LIBERO benchmark and 5 RLBench tasks, using success rate as the metric. We also compare Audio-VLA with Open-VLA-OFT~\cite{1.4openvla} and $\pi_0$~\cite{1.5pi0} on two real-world tasks. Additionally, we conduct ablation studies in both simulation and real-world settings to investigate the effectiveness of incorporating contact audio signals into VLA.

\paragraph{Simulation Experiments Results}

Simulation experiments are conducted under both standard environment conditions and domain shift scenarios to comprehensively evaluate our proposed Audio-VLA  and comparative methods.

\noindent\textbf{Performance in Standard Environment.}
In the standard environment experiment results shown in Tab.~\ref{tab:simulation_results}, Audio-VLA achieves 97.6\% average success rate on LIBERO and 55.1\% on RLBench, outperforming all comparative methods. In contact-intensive tasks RLBench Task2 and Task3, Audio-VLA outperforms the second-best method OpenVLA-OFT~\cite{oft} by 2.7\% and 10.2\% respectively.

Our proposed Audio-VLA demonstrates that acoustic perception addresses fundamental limitations of vision-only approaches in manipulation tasks, providing irreplaceable information particularly when visual perception fails to capture contact dynamics.
Through extraction of high-frequency vibrational signals and acoustic features, Audio-VLA enables perception and interpretation of contact dynamics. This continuous acoustic feedback provides essential information about contact states and object interactions, substantially improving decision-making in complex manipulation scenarios.
Audio-VLA's performance in long-horizon tasks such as LIBERO-Long validates that contact audio enables dynamic process understanding. The acoustic events—initial contact, dynamic feedback, and completion confirmation—provide temporal consistency that maintains task coherence, which proves particularly crucial for long-horizon tasks.

\noindent\textbf{Robustness Under Domain Shift.}
When desktop texture colors and lighting undergo random modifications, all methods experience significant performance degradation, yet Audio-VLA demonstrates superior robustness. Under domain shift conditions, Audio-VLA maintains 74.7\% success rate on LIBERO, and  41.5\% on RLBench surpassing all comparative methods.

Audio-VLA demonstrates superior robustness under domain shift, with only a 23.5\% performance decline on LIBERO~\cite{LIBERO}.
 The performance degradation rate on RLBench is also substantially lower. The experimental results of Audio-VLA under domain shift demonstrate that when visual information deteriorates due to environmental changes, acoustic signals provide domain-invariant physical interaction features unaffected by variations in lighting, color, or texture—characteristics that vision-only methods cannot perceive.
In contact-intensive Tasks 2 and 3, Audio-VLA maintains over 66\% performance retention, significantly outperforming comparative methods.
The performance gap reveals that in tasks requiring precise force control and continuous state monitoring, visual modality nearly loses its ability to perceive  contact states under domain shift, whereas contact audio  provide stable and continuous feedback through  vibration patterns and frequency changes that remain invariant across visual domain variations.

\paragraph{Real-World Robot Experiment Results}

Real-world experiments are conducted under both seen and unseen environmental conditions. Table~\ref{tab:real_world_results} presents the performance of Audio-VLA and comparative methods on real-world manipulation tasks, evaluated using both binary success rate and the proposed TCR metric.

\noindent\textbf{Performance in Seen Environment.}
% As shown in Tab.~\ref{tab:real_world_results}, in seen environmental conditions, Audio-VLA demonstrates substantial advantages over $\pi_0$-FAST~\cite{1.5pi0} and OpenVLA-OFT~\cite{oft} across both tasks and evaluation metrics. For success rates, Audio-VLA achieves 60\% on EAWM and 30\% on S5GO, representing threefold improvements over both OpenVLA-OFT~\cite{oft} and $\pi_0$-FAST~\cite{1.5pi0}, which achieve 20\% and 10\% success rates respectively on the two tasks. More revealing are the TCR measurements, where Audio-VLA maintains 73\% completion rate on EAWM and 72\% on S5GO, significantly outperforming OpenVLA-OFT~\cite{oft} which achieves 45\% and 34\% respectively, and $\pi_0$-FAST~\cite{1.5pi0} which achieves 34\% and 23\% respectively. The substantial gap between low success rates of 10-20\% and much higher TCR values of 23-45\% for both comparative methods indicates fundamental vision-only limitations in contact-intensive manipulation, where visual occlusion prevents accurate contact assessment. Audio-VLA's superior TCR performance demonstrates that contact audio overcomes these limitations by monitoring friction variations and pressure changes, providing real-time feedback about dynamic interaction processes imperceptible to visual perception.
As shown in Tab.~\ref{tab:real_world_results}, in seen environmental conditions, our Audio-VLA achieves threefold improvements in success rates compared to OpenVLA-OFT~\cite{oft} and $\pi_0$-FAST~\cite{1.5pi0} on both EAWM and S5GO tasks. The TCR measurements further demonstrate Audio-VLA's superiority with over 70\% completion rates, significantly exceeding the  vision-only methods.

% The performance gap between Audio-VLA and comparative methods reveals that vision-only approaches can initiate manipulation sequences yet fail at critical contact-dependent transitions where visual occlusion prevents accurate state assessment. In EAWM, vision guides the eraser to the whiteboard but cannot determine surface contact states. Similarly, in S5GO, vision cannot model the relationship between spoon insertion depth and scooped weight.
% Audio-VLA demonstrates that contact audio overcomes these limitations by capturing interaction physics through acoustic signatures. For EAWM, audio monitors friction variations during erasing; for S5GO, it captures the correlation between insertion depth and scooped weight. This acoustic perception provides feedback about dynamic processes imperceptible to vision, transforming partial attempts into successful completions.

The performance gap between Audio-VLA and baselines shows that vision-only approaches initiate manipulation sequences but fail at contact-dependent transitions where occlusion prevents state assessment. In EAWM, vision guides the eraser to the whiteboard but cannot determine surface contact; in S5GO, it cannot model spoon depth-weight relationships.
Audio-VLA overcomes these limitations through acoustic signatures of interaction physics. Audio monitors friction variations during  EAWM and captures depth-weight correlations during S5GO, providing feedback about dynamics imperceptible to vision and enabling successful task completion.

\noindent\textbf{Robustness Under Unseen Environment.}
The results of unseen environment experiments in Tab.~\ref{tab:real_world_results} demonstrate that our Audio-VLA maintains robust performance while comparative methods experience severe degradation. Audio-VLA preserves 30\% and 20\% success rates on EAWM and S5GO respectively, whereas vision-only methods approach near-zero performance. The TCR metric reveals that Audio-VLA retains over 55\% completion rates across both tasks, significantly exceeding the sub-25\% rates of comparative methods.

The robustness differential validates that contact audio provides environment-invariant physical signals. When visual changes disrupt vision-based models, acoustic signatures of contact mechanics remain consistent. In EAWM with altered marks, friction characteristics persist regardless of visual variations; in S5GO with different oatmeal properties, acoustic patterns maintain correlation with scooping quantity. These invariant features enable Audio-VLA to preserve task understanding despite environmental perturbations.
This performance retention demonstrates that multimodal perception with contact audio enhances generalization beyond visual augmentation alone. Vision-only methods rely on appearance features that vary across environments, while contact audio captures the constant underlying physics of manipulation.

\subsection{Ablation Studies}

Ablation studies on RLBench in Table~\ref{tab:ablation_sim_results} and real robot experiments in Table~\ref{tab:ablation_real_results} reveal the importance of both audio modality integration and LoRA fine-tuning for the audio encoder.

\noindent\textbf{Audio Modality Contribution.} The inferior performance of the vision-only configuration compared to the full configuration demonstrates that audio provides critical information for contact-intensive manipulation. TCR metrics in real-world experiments reveal that contact audio can accurately capture frictional variations during wiping processes and weight feedback during scooping operations, bridging the perception gap at critical contact transitions where vision alone fails to provide sufficient feedback for force control and contact state assessment during these fine-grained manipulations.

\noindent\textbf{Audio Encoder LoRA Fine-tuning Necessity.} The w/o LoRA configuration reveals that fine-tuning the audio encoder is essential for extracting task-relevant acoustic features. The pre-trained audio encoder cannot effectively process manipulation-specific sounds without LoRA~\cite{hu2022lora}. The doubling of EAWM success rates after LoRA fine-tuning~\cite{hu2022lora} and consistent improvements across both simulation and real-world experiments demonstrate that LoRA~\cite{hu2022lora} is particularly crucial for the audio encoder to learn contact audio generated during robotic manipulation processes.

\section{CONCLUSION}

This paper presents Audio-VLA, a multimodal manipulation policy that integrates acoustic perception into VLA models to overcome vision-only limitations. We develop audio-enhanced LIBERO~\cite{LIBERO} and RLBench~\cite{james2020rlbench} simulation environments and propose the TCR metric to evaluate dynamic process perception capabilities. Experimental results demonstrate that Audio-VLA achieves superior performance in both simulation environments and real-world tasks, proving the contribution of contact audio perception in overcoming visual perception limitations and providing essential contact feedback that is imperceptible to visual sensors.

\bibliography{root}  % references.bib 文件名

\begin{thebibliography}{10}

\bibitem{1.1RT2}
B.~Zitkovich, T.~Yu, S.~Xu, P.~Xu, T.~Xiao, F.~Xia, J.~Wu, P.~Wohlhart, S.~Welker, A.~Wahid, {\em et~al.}, ``Rt-2: Vision-language-action models transfer web knowledge to robotic control,'' in {\em Conference on Robot Learning}, pp.~2165--2183, PMLR, 2023.

\bibitem{1.2Palm}
D.~Driess, F.~Xia, M.~S. Sajjadi, C.~Lynch, A.~Chowdhery, A.~Wahid, J.~Tompson, Q.~Vuong, T.~Yu, W.~Huang, {\em et~al.}, ``Palm-e: An embodied multimodal language model,'' 2023.

\bibitem{1.3flamingo}
J.-B. Alayrac, J.~Donahue, P.~Luc, A.~Miech, I.~Barr, Y.~Hasson, K.~Lenc, A.~Mensch, K.~Millican, M.~Reynolds, {\em et~al.}, ``Flamingo: a visual language model for few-shot learning,'' {\em Advances in neural information processing systems}, vol.~35, pp.~23716--23736, 2022.

\bibitem{1.4openvla}
M.~J. Kim, K.~Pertsch, S.~Karamcheti, T.~Xiao, A.~Balakrishna, S.~Nair, R.~Rafailov, E.~Foster, G.~Lam, P.~Sanketi, {\em et~al.}, ``Openvla: An open-source vision-language-action model,'' {\em arXiv preprint arXiv:2406.09246}, 2024.

\bibitem{1.5pi0}
K.~Black, N.~Brown, D.~Driess, A.~Esmail, M.~Equi, C.~Finn, N.~Fusai, L.~Groom, K.~Hausman, B.~Ichter, S.~Jakubczak, T.~Jones, L.~Ke, S.~Levine, A.~Li-Bell, M.~Mothukuri, S.~Nair, K.~Pertsch, L.~X. Shi, J.~Tanner, Q.~Vuong, A.~Walling, H.~Wang, and U.~Zhilinsky, ``$\pi_0$: A vision-language-action flow model for general robot control,'' 2024.

\bibitem{deitke2020robothor}
M.~Deitke, W.~Han, A.~Herrasti, A.~Kembhavi, E.~Kolve, R.~Mottaghi, J.~Salvador, D.~Schwenk, E.~VanderBilt, M.~Wallingford, {\em et~al.}, ``Robothor: An open simulation-to-real embodied ai platform,'' in {\em Proceedings of the IEEE/CVF conference on computer vision and pattern recognition}, pp.~3164--3174, 2020.

\bibitem{o2024openx}
A.~O’Neill, A.~Rehman, A.~Maddukuri, A.~Gupta, A.~Padalkar, A.~Lee, A.~Pooley, A.~Gupta, A.~Mandlekar, A.~Jain, {\em et~al.}, ``Open x-embodiment: Robotic learning datasets and rt-x models: Open x-embodiment collaboration 0,'' in {\em 2024 IEEE International Conference on Robotics and Automation (ICRA)}, pp.~6892--6903, IEEE, 2024.

\bibitem{LIBERO}
B.~Liu, Y.~Zhu, C.~Gao, Y.~Feng, Q.~Liu, Y.~Zhu, and P.~Stone, ``Libero: Benchmarking knowledge transfer for lifelong robot learning,'' in {\em Advances in Neural Information Processing Systems} (A.~Oh, T.~Naumann, A.~Globerson, K.~Saenko, M.~Hardt, and S.~Levine, eds.), vol.~36, pp.~44776--44791, Curran Associates, Inc., 2023.

\bibitem{james2020rlbench}
S.~James, Z.~Ma, D.~R. Arrojo, and A.~J. Davison, ``Rlbench: The robot learning benchmark \& learning environment,'' {\em IEEE Robotics and Automation Letters}, vol.~5, no.~2, pp.~3019--3026, 2020.

\bibitem{mees2022calvin}
O.~Mees, L.~Hermann, E.~Rosete-Beas, and W.~Burgard, ``Calvin: A benchmark for language-conditioned policy learning for long-horizon robot manipulation tasks,'' {\em IEEE Robotics and Automation Letters}, vol.~7, no.~3, pp.~7327--7334, 2022.

\bibitem{1.6touch}
M.~Eckstein, I.~Mamaev, B.~Ditzen, and U.~Sailer, ``Calming effects of touch in human, animal, and robotic interaction—scientific state-of-the-art and technical advances,'' {\em Frontiers in psychiatry}, vol.~11, p.~555058, 2020.

\bibitem{1.6touch2}
M.~Iskandar, A.~Albu-Sch{\"a}ffer, and A.~Dietrich, ``Intrinsic sense of touch for intuitive physical human-robot interaction,'' {\em Science Robotics}, vol.~9, no.~93, p.~eadn4008, 2024.

\bibitem{liu2024maniwav}
Z.~Liu, C.~Chi, E.~Cousineau, N.~Kuppuswamy, B.~Burchfiel, and S.~Song, ``Maniwav: Learning robot manipulation from in-the-wild audio-visual data,'' {\em arXiv preprint arXiv:2406.19464}, 2024.

\bibitem{mejia2024hearing}
J.~Mejia, V.~Dean, T.~Hellebrekers, and A.~Gupta, ``Hearing touch: Audio-visual pretraining for contact-rich manipulation,'' in {\em 2024 IEEE International Conference on Robotics and Automation (ICRA)}, pp.~6912--6919, IEEE, 2024.

\bibitem{hao2025tla}
P.~Hao, C.~Zhang, D.~Li, X.~Cao, X.~Hao, S.~Cui, and S.~Wang, ``Tla: Tactile-language-action model for contact-rich manipulation,'' {\em arXiv preprint arXiv:2503.08548}, 2025.

\bibitem{zhang2025vtla}
C.~Zhang, P.~Hao, X.~Cao, X.~Hao, S.~Cui, and S.~Wang, ``Vtla: Vision-tactile-language-action model with preference learning for insertion manipulation,'' {\em arXiv preprint arXiv:2505.09577}, 2025.

\bibitem{1.3six}
M.~Wilson, ``Six views of embodied cognition,'' {\em Psychonomic bulletin \& review}, vol.~9, no.~4, pp.~625--636, 2002.

\bibitem{1.3gupta2021embodied}
A.~Gupta, S.~Savarese, S.~Ganguli, and L.~Fei-Fei, ``Embodied intelligence via learning and evolution,'' {\em Nature communications}, vol.~12, no.~1, p.~5721, 2021.

\bibitem{cheng2025omnivtla}
Z.~Cheng, Y.~Zhang, W.~Zhang, H.~Li, K.~Wang, L.~Song, and H.~Zhang, ``Omnivtla: Vision-tactile-language-action model with semantic-aligned tactile sensing,'' {\em arXiv preprint arXiv:2508.08706}, 2025.

\bibitem{van2018slip}
K.~Van~Wyk and J.~Falco, ``Slip detection: Analysis and calibration of univariate tactile signals,'' {\em arXiv preprint arXiv:1806.10451}, 2018.

\bibitem{ryun2017tactile}
S.~Ryun, J.~S. Kim, H.~Lee, and C.~K. Chung, ``Tactile frequency-specific high-gamma activities in human primary and secondary somatosensory cortices,'' {\em Scientific reports}, vol.~7, no.~1, p.~15442, 2017.

\bibitem{1.4llama2}
H.~Touvron, L.~Martin, K.~Stone, P.~Albert, A.~Almahairi, Y.~Babaei, N.~Bashlykov, S.~Batra, P.~Bhargava, S.~Bhosale, {\em et~al.}, ``Llama 2: Open foundation and fine-tuned chat models,'' {\em arXiv preprint arXiv:2307.09288}, 2023.

\bibitem{oquab2023dinov2}
M.~Oquab, T.~Darcet, T.~Moutakanni, H.~Vo, M.~Szafraniec, V.~Khalidov, P.~Fernandez, D.~Haziza, F.~Massa, A.~El-Nouby, {\em et~al.}, ``Dinov2: Learning robust visual features without supervision,'' {\em arXiv preprint arXiv:2304.07193}, 2023.

\bibitem{siglip}
X.~Zhai, B.~Mustafa, A.~Kolesnikov, and L.~Beyer, ``Sigmoid loss for language image pre-training,'' in {\em Proceedings of the IEEE/CVF international conference on computer vision}, pp.~11975--11986, 2023.

\bibitem{guzhov2022audioclip}
A.~Guzhov, F.~Raue, J.~Hees, and A.~Dengel, ``Audioclip: Extending clip to image, text and audio,'' in {\em ICASSP 2022-2022 IEEE International Conference on Acoustics, Speech and Signal Processing (ICASSP)}, pp.~976--980, IEEE, 2022.

\bibitem{hu2022lora}
E.~J. Hu, Y.~Shen, P.~Wallis, Z.~Allen-Zhu, Y.~Li, S.~Wang, L.~Wang, W.~Chen, {\em et~al.}, ``Lora: Low-rank adaptation of large language models.,'' {\em ICLR}, vol.~1, no.~2, p.~3, 2022.

\bibitem{pumacay2024colosseum}
W.~Pumacay, I.~Singh, J.~Duan, R.~Krishna, J.~Thomason, and D.~Fox, ``The colosseum: A benchmark for evaluating generalization for robotic manipulation,'' {\em arXiv preprint arXiv:2402.08191}, 2024.

\bibitem{rt1}
A.~Brohan, N.~Brown, J.~Carbajal, Y.~Chebotar, J.~Dabis, C.~Finn, K.~Gopalakrishnan, K.~Hausman, A.~Herzog, J.~Hsu, {\em et~al.}, ``Rt-1: Robotics transformer for real-world control at scale,'' {\em arXiv preprint arXiv:2212.06817}, 2022.

\bibitem{cotvla}
Q.~Zhao, Y.~Lu, M.~J. Kim, Z.~Fu, Z.~Zhang, Y.~Wu, Z.~Li, Q.~Ma, S.~Han, C.~Finn, A.~Handa, T.-Y. Lin, G.~Wetzstein, M.-Y. Liu, and D.~Xiang, ``Cot-vla: Visual chain-of-thought reasoning for vision-language-action models,'' in {\em Proceedings of the IEEE/CVF Conference on Computer Vision and Pattern Recognition (CVPR)}, pp.~1702--1713, June 2025.

\bibitem{yuan2017gelsight}
W.~Yuan, S.~Dong, and E.~H. Adelson, ``Gelsight: High-resolution robot tactile sensors for estimating geometry and force,'' {\em Sensors}, vol.~17, no.~12, p.~2762, 2017.

\bibitem{narang2021interpreting}
Y.~S. Narang, B.~Sundaralingam, K.~Van~Wyk, A.~Mousavian, and D.~Fox, ``Interpreting and predicting tactile signals for the syntouch biotac,'' {\em The International Journal of Robotics Research}, vol.~40, no.~12-14, pp.~1467--1487, 2021.

\bibitem{audioset}
J.~F. Gemmeke, D.~P. Ellis, D.~Freedman, A.~Jansen, W.~Lawrence, R.~C. Moore, M.~Plakal, and M.~Ritter, ``Audio set: An ontology and human-labeled dataset for audio events,'' in {\em 2017 IEEE international conference on acoustics, speech and signal processing (ICASSP)}, pp.~776--780, IEEE, 2017.

\bibitem{deng2009imagenet}
J.~Deng, W.~Dong, R.~Socher, L.-J. Li, K.~Li, and L.~Fei-Fei, ``Imagenet: A large-scale hierarchical image database,'' in {\em 2009 IEEE conference on computer vision and pattern recognition}, pp.~248--255, Ieee, 2009.

\bibitem{clip}
A.~Radford, J.~W. Kim, C.~Hallacy, A.~Ramesh, G.~Goh, S.~Agarwal, G.~Sastry, A.~Askell, P.~Mishkin, J.~Clark, {\em et~al.}, ``Learning transferable visual models from natural language supervision,'' in {\em International conference on machine learning}, pp.~8748--8763, PmLR, 2021.

\bibitem{alamri2019audio}
H.~Alamri, V.~Cartillier, A.~Das, J.~Wang, A.~Cherian, I.~Essa, D.~Batra, T.~K. Marks, C.~Hori, P.~Anderson, {\em et~al.}, ``Audio visual scene-aware dialog,'' in {\em Proceedings of the IEEE/CVF Conference on Computer Vision and Pattern Recognition}, pp.~7558--7567, 2019.

\bibitem{peract}
M.~Shridhar, L.~Manuelli, and D.~Fox, ``Perceiver-actor: A multi-task transformer for robotic manipulation,'' in {\em Conference on Robot Learning}, pp.~785--799, PMLR, 2023.

\bibitem{fu2024mobile}
Z.~Fu, T.~Z. Zhao, and C.~Finn, ``Mobile aloha: Learning bimanual mobile manipulation with low-cost whole-body teleoperation,'' {\em arXiv preprint arXiv:2401.02117}, 2024.

\bibitem{quigley2009ros}
M.~Quigley, K.~Conley, B.~Gerkey, J.~Faust, T.~Foote, J.~Leibs, R.~Wheeler, A.~Y. Ng, {\em et~al.}, ``Ros: an open-source robot operating system,'' in {\em ICRA workshop on open source software}, vol.~3, p.~5, Kobe, 2009.

\bibitem{oft}
M.~J. Kim, C.~Finn, and P.~Liang, ``Fine-tuning vision-language-action models: Optimizing speed and success,'' {\em arXiv preprint arXiv:2502.19645}, 2025.

\end{thebibliography}
\bibliographystyle{ieeetr}  % 或其他样式如 plain, alpha, unsrt

\end{document}